\documentclass[review]{elsarticle}
\usepackage{hyperref}
\usepackage{cite}
\usepackage{amsmath,amssymb,amsfonts}
\usepackage{algorithmic}
\usepackage{graphicx}
\usepackage{textcomp}
\usepackage{caption}
\usepackage{subcaption}
\usepackage{multirow}
\usepackage{xcolor}

\journal{Expert Systems with Applications}

\biboptions{authoryear}

\begin{document}

\begin{frontmatter}

\title{Clickbait Headline Detection in Indonesian News Sites using Multilingual Bidirectional Encoder Representations from Transformers (M-BERT)}


\author{Muhammad Noor Fakhruzzaman \corref{mycorrespondingauthor}}
\ead{ruzza@stmm.unair.ac.id}

\author{Sa'idah Zahrotul Jannah}
\ead{saidahzj@stmm.unair.ac.id}

\author{Ratih Ardiati Ningrum}
\ead{ratih.an@stmm.unair.ac.id}

\author{Indah Fahmiyah}
\ead{indah.fahmiyah@stmm.unair.ac.id}


\cortext[mycorrespondingauthor]{Corresponding author (tel. +62811868567)}

\address{Faculty of Advanced Technology and Multidiscipline}
\address{Universitas Airlangga}
\address{Surabaya, Indonesia}

\begin{abstract}
Click counts are related to the amount of money that online advertisers paid to news sites. Such business models forced some news sites to employ a dirty trick of click-baiting, i.e., using a hyperbolic and interesting words, sometimes unfinished sentence in a headline to purposefully tease the readers. Some Indonesian online news sites also joined the party of clickbait, which indirectly degrade other established news sites' credibility. A neural network with a pre-trained language model M-BERT that acted as a embedding layer is then combined with a 100 nodes hidden layer and topped with a sigmoid classifier was trained to detect clickbait headlines. With a total of 6632 headlines as a training dataset, the classifier performed remarkably well. Evaluated with 5-fold cross validation, it has an accuracy score of 0.914, f1-score of 0.914, precision score of 0.916, and ROC-AUC of 0.92. The usage of multilingual BERT in Indonesian text classification task was tested and is possible to be enhanced further. Future possibilities, societal impact, and limitations of the clickbait detection are discussed. 
\end{abstract}


\end{frontmatter}


\section{Introduction}

Journalism has changed. Before the emergence of internet news, we bought newspapers because we were enticed to the headline on the front page which usually leads to the truth, but not anymore. The emergence of online news outlets created a whole new scheme of making money in the journalism world, the online ad. With internet advertising, a single click means money, even though it is not as much as a newspaper sale or advertising money from the sponsors, like the olden days. Now, a post headline has to rake in engagement, the metric that measures ratings in online news. 

The scheme of online advertising that bases on engagement have a negative influence on the original journalism idea. Sadly, online news organization now hunts for click money instead of the truth \citep{chen_news_2015}. 

This phenomenon promotes a unique style of headline writing, infamously known as clickbait. The more people click the post, the more engagement that post has, the more advertising value the site will gain. A study found that most of the online news organization relies on clickbait's ad money to support their daily activities. With an increasing number of online news sites in recent years, they have to contest for reader's clicks \citep{chakraborty_stop_2016}. What makes it worse is that some news source that is once credible are also retreating to the means of clickbaiting. Further obscuring the integrity of the Indonesian online news organization, as a previous study found that the usage of clickbait worsens the news site reputation \citep{hurst_clickbait_2016}.

Clickbait refers to a headline sentence that contains hyperbolic words to persuade its reader to click the following link but mostly did not reveal any major information. It may also contain a message that is controversial but did not disclose complete information about it in the sentence \citep{potthast_clickbait_2016}. Some clickbait headlines often use trending buzzwords, but most of its following link leads to complete misunderstanding.

\begin{figure}[!htb]
\center{\includegraphics[width=.7\columnwidth]
{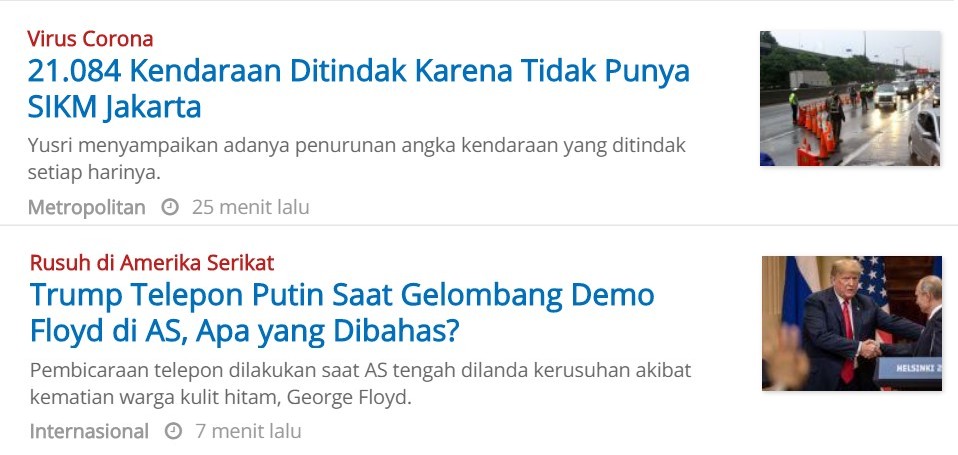}}
\caption{\label{headlineexample} An example of Clickbait and Non-Clickbait headline}
\end{figure}

An example of clickbait and non-clickbait headline is depicted in Figure \ref{headlineexample}. The first headline, which is a non-clickbait headline, translates to "21.084 Vehicles Ticketed Due to Inability to Show SIKM Jakarta". While the second headline, which is a clickbait headline, translates to "Trump Calls Putin While George Floyd Protest Enrages in the USA, What Did They Talked About?". The example shows a clear distinction of non-clickbait headline versus a clickbait headline. A non-clickbait headline delivers straightforward key information while clickbait headline entices us to seek more.

The usage of clickbait capitalized on the human's nature of curiosity. Such curiosity arises when human wants to know about something new, that they feel the gap between something they already know and something they want to know \citep{loewenstein_psychology_1994}. That curiosity gap is exploited by providing teaser messages in clickbait headline which then signals the reader about new information, provoking the reader's curiosity, and leads them to click the headline \citep{anand_we_2017}.

Previous studies about automatic clickbait detection used a neural network that was trained on a specific labeled corpus of clickbait. Past researchers also tried to find a specific pattern in their corpus of clickbaits, but the pattern is constantly changing overtime \citep{anand_we_2017,agrawal_clickbait_2016}. 

The need for automatic clickbait detection were addressed on previous studies, but mostly trained on English clickbait corpus \citep{anand_we_2017,chakraborty_stop_2016,potthast_clickbait_2016}. As \citet{zuhroh_clickbait_2020} stated in their literature review, there is a gap to be filled in Indonesian clickbait detection. Indonesia needs such a tool to increase the quality of the journalism itself, while also indirectly enhance the public digital literacy as the use of clickbaits in Indonesian online news enrages.
A past study that focused on detecting Indonesian clickbait using neural network utilized TF-IDF as their feature extraction algorithm. The TF-IDF (Term Frequency - Inverse Document Frequency) algorithm represented the feature of a text by counting the term or word appearance frequency in a document to express its relevance in a corpus \citep{maulidi_penerapan_2018}.

However, term frequency is simply not enough to capture the characteristics of clickbait headlines. In order to capture semantic and syntactic properties in the headline, the text in this study is represented with word embeddings. Word embeddings is a text feature extraction technique that maps the words into a vector space model, thus representing each word as vector and enables computers to measure distances between words, thus returning word similarity \citep{anand_we_2017,zuhroh_clickbait_2020}.

Recently, a state-of-the-art language representation model was released, named BERT. It performed the best among the available language models in completing NLP tasks \citep{devlin_bert_2018}. With the availability of the trained language model in Indonesian and other languages, this study used a pre-trained multilingual BERT model as the language model. The use of transfer learning from the multilingual model enables the model to extract the features from some headlines that used both English and Indonesian in its sentence.    

The approach of using neural network to classify clickbait was deemed to be feasible due to the dynamic nature of clickbait writing \citep{zuhroh_clickbait_2020,maulidi_penerapan_2018,agrawal_clickbait_2016}. Moreover, \citet{anand_we_2017} compared their neural network performance in detecting clickbait to other baseline model --i.e., support vector machine, decision tree, and random forest. Their neural network performed with higher accuracy for clickbait detection. 

Therefore, this study attempted to use M-BERT as an embedding layer in a neural network to detect clickbait in Indonesian online news site.

\section{Methods}
\begin{figure}[!htb]
\center{\includegraphics[width=0.8\columnwidth]
{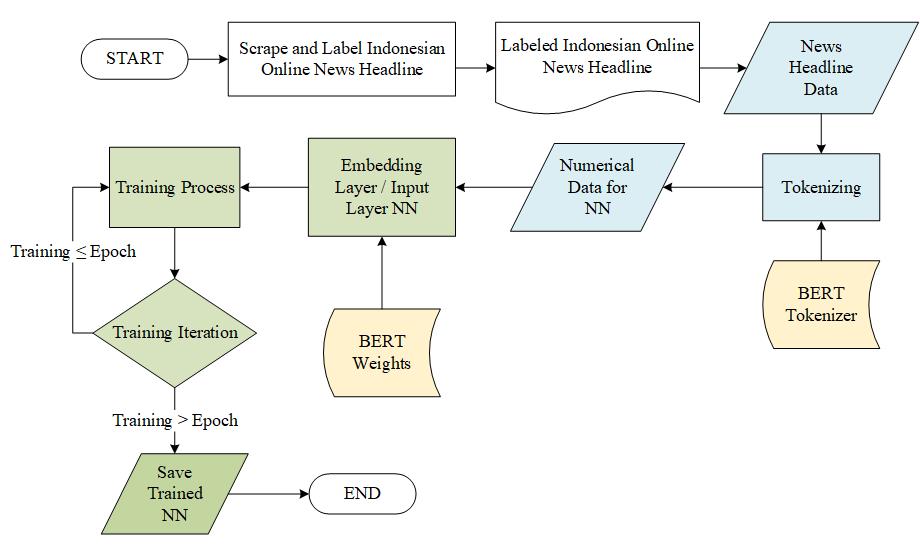}}
\caption{\label{flowtrain} Flowchart of Neural Network Training Process}
\end{figure}

The training process of the neural network used in this study is depicted in the flowchart on Figure \ref{flowtrain}. 

BERT Weights from the flowchart refers to the pre-trained model that is used for an embedding layer. While BERT tokenizer is a built-in tokenizing factory that split the strings of headline into tokens.

\subsection{Data Preparation}
The news headline corpus was retrieved from \citet{william_click-id_2020}'s dataset, consisting of 8613 annotated news headline from 12 online news sites. The news sites (partially redacted) are listed on Table \ref{scrapetab}.

\begin{table}
\caption{List of News Sites in the Dataset} \label{scrapetab}
\centering
\begin{tabular}{|c|c|}
\hline

No. & Site's URL\\    \hline
1 & http://www.de**k.com/\\ \hline 
2 & http://www.trib**news.com/\\ \hline
3 & http://www.pos**tro.com/\\ \hline
4 & http://www.repu**ika.co.id/\\ \hline
5 & http://www.ka***lagi.com/\\ \hline
6 & http://www.k*mp*s.com/\\ \hline
7 & http://www.te*po.co/\\ \hline
8 & http://www.ok**one.com/\\ \hline
9 & http://www.fim**a.com/\\ \hline
10 & http://www.lip**an6.com/\\ \hline
11 & http://sin***ews.com/\\ \hline
12 & http://www.w**ke*en.com/\\ \hline

\end{tabular}
\end{table}

The dataset contains 15,000 headlines. They were labeled by 3 undergraduate students per headline and then was deemed moderately reliable with Fleiss' Kappa Interrater agreement of 0.42 \citep{william_click-id_2020}. 

However, in this study, only the headlines which every rater agreed that it is a clickbait, are selected. With that, the dataset is now consisted of 8613 headlines, with Fleiss' Kappa of 1 which means full agreement between raters. Hence, the dataset was deemed strongly reliable. The used dataset was available online for reproducibility \citep{william_click-id_2020}. 

To explain further and make a clear distinction between clickbait and non-clickbait. Various sources listed the criteria of clickbait headlines \citep{potthast_crowdsourcing_2018,biyani__2016}. The criteria are listed on Table \ref{criteriatab} 

\begin{table}
\caption{List of Clickbait Criteria} \label{criteriatab}
\centering
\begin{tabular}{|c|c|}
\hline

No. & Criterion\\    \hline
1 & Contains teaser message to entice curiosity\\ \hline 
2 & Contains following images or video\\ \hline
3 & Contains controversial word and/or buzzword\\ \hline
4 & Contains hyperbolic sentence and/or word\\ \hline
5 & Contains question for the reader\\ \hline
6 & Contains emotion-provoking word\\ \hline

\end{tabular}
\end{table}


\subsection{Data Preprocessing}

The dataset loaded directly to the python code which is also available on the github link specified in the end of this article. Because the class is imbalanced, the data was normalized. 3316 non-clickbait headlines were random-picked to balance the dataset. The clickbait headlines still consist of 3316 data. The total counts of data for training was 6632 headlines. 

Then, using the BERT Tokenizer from huggingface, the text is stemmed, tokenized into words, padded, and indexed while also formatted accordingly for BERT layer input specification. The texts were then have its stopwords removed by using an opensource Indonesian stopword remover PySastrawi \citep{robbani_github_2018, wolf_transformers_2019}. 

Finally, each headline in the dataset is transformed into list of sequences of token ids. The sequence of integers, which refers to respective words on the dictionary, is ready to be fed to the neural network \citep{wolf_transformers_2019}. 


\subsection{Neural Network Configuration}
The neural network configuration is as follows. The input layers are two Keras input layers, each responsible for handling a list of sequences of token ids, and attention masks (the marker for pad and non-pad tokens), it is then passed forward to the BERT layer. The embedding layer is a BERT multilingual model that trained on a multilingual Wikipedia dump dataset, which included the Indonesian language \citep{devlin_bert_2018}.
The usage of a pre-trained language model allows the researcher to capture semantics features in the headline, it also enables the researcher to extract features from the headline corpus in a short time, without wasting a lot of computing hours to train a language model. 

The hidden layer consisted of 100 densely connected neurons, activated with the ReLU function. Before the sequence is passed into the Dense layer, it passes through a GlobalAveragePooling layer and Flattened to fit the dense layer's dimension. 

\begin{figure}[!htb]
\center{\includegraphics[width=0.6\columnwidth]
{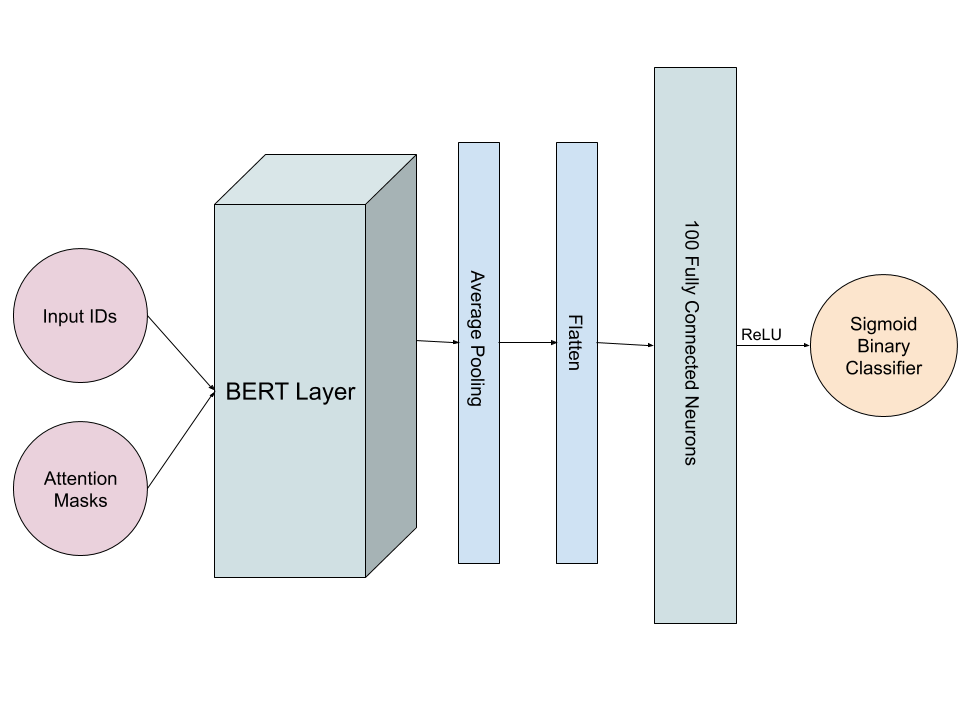}}
\caption{\label{netw} Neural Network Configuration}
\end{figure}

Finally, the sequence is passed into a final dense layer, activated with a sigmoid function to classify it as either clickbait class or non-clickbait class. The model is compiled with Adam Optimizer and a learning rate of 1e-05, then trained in three iterations. The network architecture is depicted on Figure \ref{netw}. 

\subsection{Model Evaluation}
 The model was evaluated using 5-fold cross-validation method to identify its accuracy, confusion matrices, and its ROC-AUC plot. An additional evaluation was also employed. A total of 3237 labeled headlines collected on May 2020, which is different than the training dataset collection time, was used as a test data. It is then evaluated using the same metric. The additional evaluation was employed to test whether or not the model can detect clickbaits in another dataset with different topics and possibly different clickbait sentence structure.

\section{Results}
A total of 6632 headlines were used as a training data. Specifically, 3316 headlines were labeled as Clickbait, and 3316 headlines were random-picked from a total of 5297 Non-Clickbait headlines to balance the class. 

\subsection{Exploratory Data Analysis}
\begin{figure}[!htbp]
  \begin{subfigure}[b]{.5\columnwidth}
    
    \center{\includegraphics[width=\columnwidth]{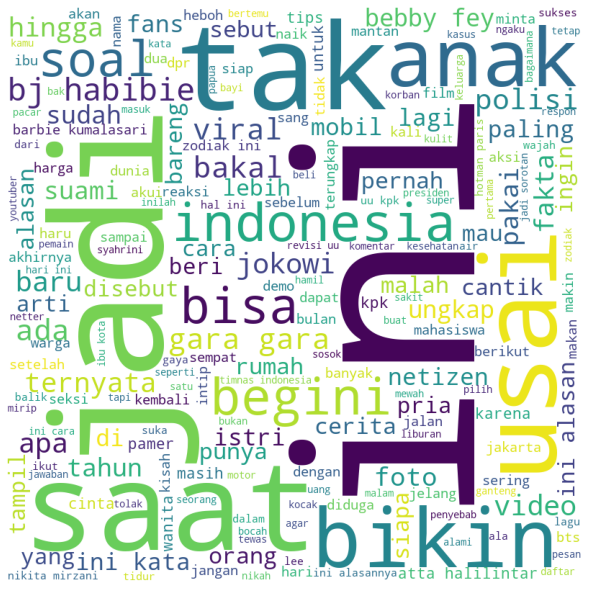}}
    
    \caption{Clickbait}
    \label{wordcloudclick}
  \end{subfigure}%
  \begin{subfigure}[b]{.5\columnwidth}
    \center{\includegraphics[width=\columnwidth]{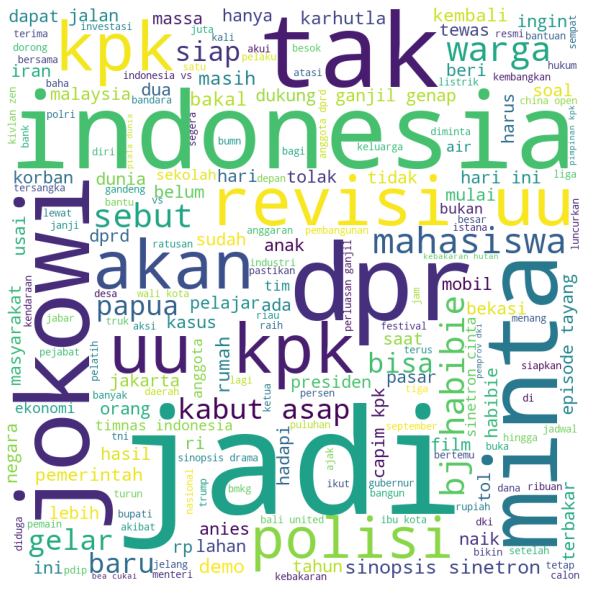}}
    
    \subcaption{Non-clickbait}
    \label{wordcloudnon}
  \end{subfigure}
  \caption{Word clouds of Clickbait and Non-Clickbait}
  \label{wordclouds}
\end{figure}

In the clickbait category, the word "ini" or "this" in English, appears highly frequently, as seen in Figure \ref{wordcloudclick}. Usually, the word "ini" is used as a pointing word that leads the reader to the curiosity gap e.g. "This 5 Kinds Fruit is Really Good For Your Skin!" which is translated from Indonesian sentence "5 Macam Buah Ini Sangat Bagus Untuk Kulitmu!". Notice the use of "ini" in the headline.

Although it may be just one signal word of clickbait, but it seems that clickbaiting uses this word very often. Hence, it appears as a top word in the clickbait category.

Additionally, the word "bikin" is also appearing frequently in the clickbait category. The word "bikin" is considered a conversational slang in Indonesia, mostly used among urban citizen. It is well-suited for a clickbait because slangs are often used in clickbait headline to bring the headline to an "easier level", so that the readers can relate to the headline with relative ease 

While the word "jadi" can mean two different things. One of it means "become" if it is used in a less formal setting, it can also be used as a conjunction word, often coupled with other sentence, which then the word "jadi" translates to "so" in English. 

The word "jadi" appears oftenly both in the clickbait category and non-clickbait category, this may confuse the model, because the word has different meaning in Indonesian. However, using M-BERT can solve this confusion because M-BERT can capture semantic meaning by the context of the sentence it appeared in.

Other words in the wordcloud depicted on Figure \ref{wordclouds} also shows some hints on which topic often appeared in clickbait category. Celebrity names appeared in the clickbait wordcloud which indicated that clickbaits are often used in gossip and tabloid headlines, or "soft news". Still, some "hard news" words are also appearing often, such as "jokowi","bj habibie", and "indonesia". Which indicated that clickbaits are used among the "hard news" topics as well.

Meanwhile, the non-clickbait wordcloud depicted in Figure \ref{wordcloudnon} shows a lot of "hard news" related words e.g. "indonesia", "karhutla", "kpk","polisi" and no signs of neither celebrity names nor informal words. 

\begin{figure}[!htbp]
  \begin{subfigure}[b]{.5\columnwidth}
    
    \center{\includegraphics[width=\columnwidth]{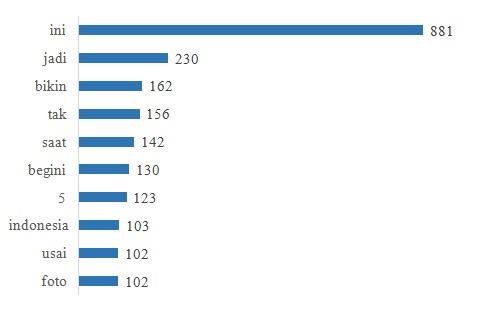}}
    
    \caption{Clickbait}
    \label{barclick}
  \end{subfigure}%
  \begin{subfigure}[b]{.5\columnwidth}
    \center{\includegraphics[width=\columnwidth]{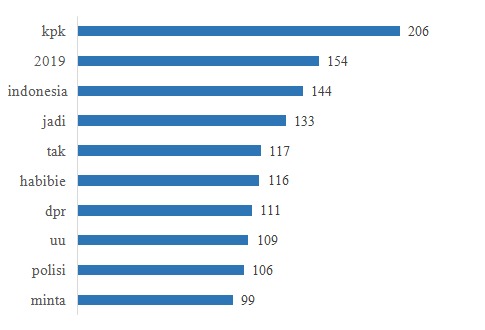}}
    
    \subcaption{Non-clickbait}
    \label{barnon}
  \end{subfigure}
  \caption{Top 10 Words of Clickbait and Non-Clickbait}
  \label{bartop}
\end{figure}

Top 10 words can also describe the corpus. Using a bag of words tokenization, bar charts depicting top 10 appearing words in the corpus on each category are depicted on Figure \ref{bartop}.

Looking at Figure \ref{barclick}, the word "ini" appeared 881 times in the corpus, which is a lot more often compared to other words. Also, most of the top 10 words in clickbait did not refer directly to the topic of the article, although a good headline should refer the topic directly. Clickbait bar chart shows a good description of the corpus and match the clickbait criteria well.

Whereas on Figure \ref{barnon}, "kpk" appeared most often with 206 frequency. Looking at all the top 10 words of non-clickbait corpus, most of them were nouns, e.g. "Indonesia", "habibie", "dpr". It shows that non-clickbait headline often refer to a specific topic directly, without using pointing words or conjunctions, unlike clickbait. 

From the word clouds and bar charts, the distinction between two categories lies on the use of informal words, named entities, and different parts-of-speech usage.

\subsection{Initial Model Evaluation}
After the dataset is trained on the described neural network model for 3 iterations, it is evaluated using 5-fold cross-validation, with each fold yielded accuracy, precision, recall, and f1-score, which are then used for evaluating the model.

Accuracy, precision, recall, and f1-score values indicate the performance of the model, the closer the value is to 1, the better the model in classifying headlines. 

Specifically, accuracy represents the model's ability to classify each headline into their supposed class accurately. Then, precision represents the proportion of true positives i.e., predicted as clickbait turned out true, among the total predicted clickbait class.

Furthermore, recall represents the proportion of true positives related to the actual clickbait class. While f1 score is a combination of precision and recall that can represent both values in equal proportion. Table \ref{modelresults} shows evaluation metrics from all folds in the cross-validation. 

\begin{table}[!htb]
\caption{Performance Evaluation Metrics} \label{modelresults}
\centering
\begin{tabular}{|c|c|c|c|c|}
\hline

\textbf{Fold} & \textbf{Accuracy} & \textbf{Precision} & \textbf{Recall} & \textbf{f1-score}\\    \hline
1 & 0.89 & 0.90 & 0.89 & 0.89\\ \hline
2 & 0.93 & 0.93 & 0.93 & 0.93\\ \hline
3 & 0.92 & 0.92 & 0.92 & 0.92\\ \hline
4 & 0.91 & 0.91 & 0.91 & 0.91\\ \hline
5 & 0.92 & 0.92 & 0.92 & 0.92\\ \hline

\end{tabular}
\end{table}

\begin{figure}[!htb]
\center{\includegraphics[width=0.7\columnwidth]
{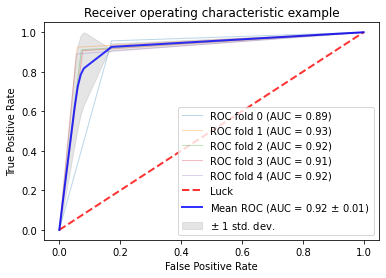}}
\caption{\label{rocauc} ROC-AUC Plot}
\end{figure}

Then, the ROC (Receiver Operating Characteristic) curve and AUC (Area Under the Curve) is calculated to further evaluate the model. The ROC curve shows the relation between the True Positive Rate and False Positive Rate. Ideally, the model is expected to maximize the True Positive Rate and minimize the False Positive Rate. Wider AUC with a score closer to 1 is considered a better model. 

Figure \ref{rocauc} depicts that the ROC curve of each fold is close to the y-axis, followed by a decent AUC score with an average of 0.92($\pm$0.01).

Furthermore, other classifier, i.e. Bidirectional LongShort Term Memory (Bi-LSTM) and Convolutional Neural Network (CNN) using \citet{chakraborty_stop_2016}'s configuration, also an XGBoost classifier with TF-IDF vectors, were used as a comparison. The performance of those classifiers compared to this study M-BERT based model is depicted on Table \ref{compared}. The table showed that M-BERT performed better compared to Bi-LSTM, CNN, and XGBoost in this task.

\begin{table}[!htb]
\caption{Performance Comparison} \label{compared}
\centering
\begin{tabular}{|c|c|}
\hline

\textbf{Model Name} & \textbf{Average Accuracy} \\    \hline
M-BERT & 0.9153 \\ \hline
Bi-LSTM & 0.8125 \\ \hline
CNN & 0.7958 \\ \hline
XGBoost & 0.8069 \\ \hline

\end{tabular}
\end{table}

\subsection{Additional Model Evaluation}
Additionally, a further evaluation is employed to see whether or not the model can adapt to a dataset from different collection time period with different media agenda. Specifically, the training dataset is collected pre-Covid19 while the additional test dataset is collected during Covid19 pandemic. 

\begin{figure}[!ht]
  \begin{subfigure}[b]{.5\columnwidth}
    
    \center{\includegraphics[width=\columnwidth]{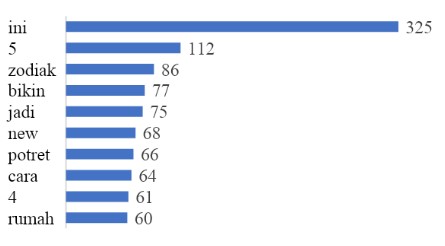}}
    
    \caption{Clickbait}
    \label{wordfreqclick}
  \end{subfigure}%
  \begin{subfigure}[b]{.5\columnwidth}
    \center{\includegraphics[width=\columnwidth]{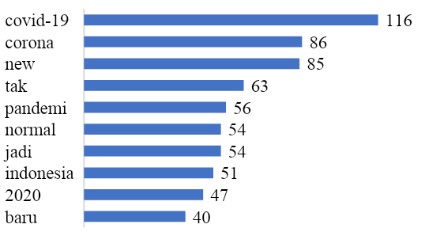}}
    
    \subcaption{Non-clickbait}
    \label{wordfreqnon}
  \end{subfigure}
  \caption{Top 10 Word frequencies of Additional Test Data}
  \label{wordfreq}
\end{figure}

A total of 3237 new annotated headlines from May 2020 were pre-processed through the same method and then fed to the model to classify, then evaluated with the ground truth. The word frequency of the data for additional evaluation is depicted on Figure \ref{wordfreq}

Depicted in Figure \ref{wordfreqclick}, clickbait headlines still has the word "ini" mentioned quite frequently, which is similar to the clickbait category in the training dataset. It was mentioned 325 times in the clickbait group.

However, in Figure \ref{wordfreqnon}, "Covid-19" is the most frequent word among the non-clickbait groups, mentioned 116 times. Other top 10 frequent words are still related to "Covid-19", i.e. "corona", "new", "normal", and "pandemi", which is consistent with the media agenda-setting and public issue during the data acquisition. These difference of agenda may confuse the model, and the additional evaluation seeks to identify those flaws.

The additional evaluation shows average accuracy of 0.83, precision of 0.82, recall of 0.83, and f1-score of 0.83. 

The additional evaluation score indicated that the model can still detect newer headline, given different topic, and term frequency. Although, with decreased performance.

Finally, considering all of the evaluation metrics, ROC-AUC scores, and the additional evaluation, the model was deemed to perform well.

\section{Discussions}
The finding indicates a possible future in using a pre-trained language model in classifying clickbait. With such a carefully curated dataset, the clickbait detection can be further expanded to different NLP tasks as well \citep{william_click-id_2020}.
By using BERT, the whole model looks simplified, using only a BERT layer and a hidden standard dense layer, finally topped with a sigmoid activated neuron, the classifier worked remarkably well with an average accuracy of 92\%.

However, with additional evaluation using newer dataset, the model performance decreased. It may be due to the different main issue in the new dataset. A further study is needed to evaluate the model versatility. Moreover, training a Neural Network with M-BERT took a lot of computing resource. If efficiency is the priority, XGBoost can perform moderately well (80\% avg.).

This study adds to the body of research in mass communication studies by providing an alternative tool in detecting clickbait in online news headlines, specifically in Indonesian. There are other methods, however, both quantitatively and qualitatively that scientifically assess clickbait detection, mostly in linguistic and mass communication studies.  

Indeed, this study chose to contribute from the perspective of natural language processing and expanding the body of research in applied machine learning by implementing a neural network in Indonesian mass media domain. Future researchers may delve deeper into the area of applied machine learning for mass media by looking into other methods or optimizing the current method.

The presence of clickbait detection may help the public to increase digital literacy by giving extra attention to the flagged headline so that they know what specific characteristics a clickbait headline has. Likewise, it might be able to discourage some news sites in using clickbait and shift their practices slowly into more high-quality reporting. Although, it seems unlikely due to the current funding scheme in the online advertisement domain. 

Furthermore, if the model is deployed into usable software, it can help to alleviate the bias given by the priming words that are widely used in writing clickbait headlines. Flagging the clickbait headline gives a chance for the reader to rethink their urge to click and fill their curiosity gap \citep{loewenstein_psychology_1994}.

\subsection{Limitation and Future Research}
This study has some limitations, one of which is the broad topics of news in the dataset. Such broad topic may provide noise, because clickbait headlines often dominated the celeb and gossip topic. It may affect future prediction due to the noise provided by the mentioned topic. 

Future research may fill the gap by focusing on specific hard news topic, such as politics, to learn further about clickbait usage in different topics.

Additionally, the labelling of clickbait is relatively hard, even for human. Future research may expand the dataset and select only the data where every rater agreed, so that there is a clear distinction of clickbait and non-clickbait headline. Also, the initial briefing of the rater should be conducted sistematically, so that the dataset is reliable and unambiguous.

The training dataset also needed expansions by adding more data from multiple time period of collection. This may increase the model versatility by enabling it to capture the dynamic pattern of clickbait structure in various time period. 

Furthermore, This study used pre-trained BERT using  Indonesian Wikipedia dumps as well as other 100 languages, which may not contain sensational wording commonly used in clickbait headlines, due to Wikipedia writing rules. Therefore, future researchers may collect a bigger Indonesian corpus that includes offensive and rude words, slangs, and sensational words to fine-tune the BERT model, which might increase the performance of the model.

Yet, a qualitative approach regarding clickbait assessment is also needed to define clickbait characteristics thoroughly. With a detailed specification of clickbait headline, the labeling process of headlines can be less biased. Qualitative study can also confirm the descriptive analysis of this study which stated that clickbait headline often used non-topic related words and teasing words.

\section{Conclusion}
Using the neural network in classifying clickbait has been fairly common in the English language, but not in Indonesian. This study contributes to show that Multilingual BERT, a state-of-the-art model is able to classify Indonesian clickbait headlines.


Furthermore, we would like to explore more about the methods in detecting clickbait of online news headlines. We also want to deploy the model into a usable component, like a browser extension program, so that the clickbait detector can be used by the public.

Finally, future researchers can look into the effect of the clickbait detector among the general public. Whether or not it influences adult literacy and its ability to inhibit the spread of misinformation.

\section*{Additional Resource}
The complete Python notebook and datasets are stored on https://github.com/ruzcmc/ClickbaitIndo-textclassifier. 
\section*{Acknowledgment}
Training dataset provided by A. William and Y. Sari as cited. Additional test dataset provided by Fakhruzzaman, et.al.


\bibliography{clickbait-textclassify.bib}

\begin{thebibliography}{15}
\expandafter\ifx\csname natexlab\endcsname\relax\def\natexlab#1{#1}\fi
\providecommand{\url}[1]{\texttt{#1}}
\providecommand{\href}[2]{#2}
\providecommand{\path}[1]{#1}
\providecommand{\DOIprefix}{doi:}
\providecommand{\ArXivprefix}{arXiv:}
\providecommand{\URLprefix}{URL: }
\providecommand{\Pubmedprefix}{pmid:}
\providecommand{\doi}[1]{\href{http://dx.doi.org/#1}{\path{#1}}}
\providecommand{\Pubmed}[1]{\href{pmid:#1}{\path{#1}}}
\providecommand{\bibinfo}[2]{#2}
\ifx\xfnm\relax \def\xfnm[#1]{\unskip,\space#1}\fi
\bibitem[{Agrawal(2016)}]{agrawal_clickbait_2016}
\bibinfo{author}{Agrawal, A.} (\bibinfo{year}{2016}).
\newblock \bibinfo{title}{Clickbait detection using deep learning}.
\newblock In {\it \bibinfo{booktitle}{2016 2nd {International} {Conference} on
  {Next} {Generation} {Computing} {Technologies} ({NGCT})}\/} (pp.
  \bibinfo{pages}{268--272}).
\newblock \bibinfo{publisher}{IEEE}.
\bibitem[{Anand et~al.(2017)Anand, Chakraborty \& Park}]{anand_we_2017}
\bibinfo{author}{Anand, A.}, \bibinfo{author}{Chakraborty, T.}, \&
  \bibinfo{author}{Park, N.} (\bibinfo{year}{2017}).
\newblock \bibinfo{title}{We used neural networks to detect clickbaits: {You}
  won’t believe what happened next!}
\newblock In {\it \bibinfo{booktitle}{European {Conference} on {Information}
  {Retrieval}}\/} (pp. \bibinfo{pages}{541--547}).
\newblock \bibinfo{publisher}{Springer}.
\bibitem[{Biyani et~al.(2016)Biyani, Tsioutsiouliklis \&
  Blackmer}]{biyani__2016}
\bibinfo{author}{Biyani, P.}, \bibinfo{author}{Tsioutsiouliklis, K.}, \&
  \bibinfo{author}{Blackmer, J.} (\bibinfo{year}{2016}).
\newblock \bibinfo{title}{" 8 {Amazing} {Secrets} for {Getting} {More}
  {Clicks}": {Detecting} {Clickbaits} in {News} {Streams} {Using} {Article}
  {Informality}}.
\newblock In {\it \bibinfo{booktitle}{Thirtieth {AAAI} {Conference} on
  {Artificial} {Intelligence}}\/}.
\bibitem[{Chakraborty et~al.(2016)Chakraborty, Paranjape, Kakarla \&
  Ganguly}]{chakraborty_stop_2016}
\bibinfo{author}{Chakraborty, A.}, \bibinfo{author}{Paranjape, B.},
  \bibinfo{author}{Kakarla, S.}, \& \bibinfo{author}{Ganguly, N.}
  (\bibinfo{year}{2016}).
\newblock \bibinfo{title}{Stop clickbait: {Detecting} and preventing clickbaits
  in online news media}.
\newblock In {\it \bibinfo{booktitle}{2016 {IEEE}/{ACM} {International}
  {Conference} on {Advances} in {Social} {Networks} {Analysis} and {Mining}
  ({ASONAM})}\/} (pp. \bibinfo{pages}{9--16}).
\newblock \bibinfo{publisher}{IEEE}.
\bibitem[{Chen et~al.(2015)Chen, Conroy \& Rubin}]{chen_news_2015}
\bibinfo{author}{Chen, Y.}, \bibinfo{author}{Conroy, N.~J.}, \&
  \bibinfo{author}{Rubin, V.~L.} (\bibinfo{year}{2015}).
\newblock \bibinfo{title}{News in an online world: {The} need for an
  “automatic crap detector”}.
\newblock {\it \bibinfo{journal}{Proceedings of the Association for Information
  Science and Technology}\/},  {\it \bibinfo{volume}{52}\/},
  \bibinfo{pages}{1--4}.
\newblock \bibinfo{note}{ISBN: 2373-9231 Publisher: Wiley Online Library}.
\bibitem[{Devlin et~al.(2018)Devlin, Chang, Lee \&
  Toutanova}]{devlin_bert_2018}
\bibinfo{author}{Devlin, J.}, \bibinfo{author}{Chang, M.-W.},
  \bibinfo{author}{Lee, K.}, \& \bibinfo{author}{Toutanova, K.}
  (\bibinfo{year}{2018}).
\newblock \bibinfo{title}{Bert: {Pre}-training of deep bidirectional
  transformers for language understanding}.
\newblock {\it \bibinfo{journal}{arXiv preprint arXiv:1810.04805}\/}, .
\bibitem[{Hurst(2016)}]{hurst_clickbait_2016}
\bibinfo{author}{Hurst, N.} (\bibinfo{year}{2016}).
\newblock {\it \bibinfo{title}{To clickbait or not to clickbait? an examination
  of clickbait headline effects on source credibility}\/}.
\newblock Ph.D. thesis University of Missouri--Columbia.
\bibitem[{Loewenstein(1994)}]{loewenstein_psychology_1994}
\bibinfo{author}{Loewenstein, G.} (\bibinfo{year}{1994}).
\newblock \bibinfo{title}{The psychology of curiosity: {A} review and
  reinterpretation.}
\newblock {\it \bibinfo{journal}{Psychological bulletin}\/},  {\it
  \bibinfo{volume}{116}\/}, \bibinfo{pages}{75}.
\newblock \bibinfo{note}{ISBN: 1939-1455 Publisher: American Psychological
  Association}.
\bibitem[{Maulidi et~al.(2018)Maulidi, Ayilillahi, Isyiriyah \&
  Palandi}]{maulidi_penerapan_2018}
\bibinfo{author}{Maulidi, R.}, \bibinfo{author}{Ayilillahi, M.~F.},
  \bibinfo{author}{Isyiriyah, L.}, \& \bibinfo{author}{Palandi, J.~F.}
  (\bibinfo{year}{2018}).
\newblock \bibinfo{title}{Penerapan {Neural} {Network} {Backprogpagation}
  {Untuk} {Klasifikasi} {Artikel} {Clickbait}}.
\newblock In {\it \bibinfo{booktitle}{Seminar {Nasional} {FST} 2018}\/}.
\bibitem[{Potthast et~al.(2018)Potthast, Gollub, Komlossy, Schuster, Wiegmann,
  Fernandez, Hagen \& Stein}]{potthast_crowdsourcing_2018}
\bibinfo{author}{Potthast, M.}, \bibinfo{author}{Gollub, T.},
  \bibinfo{author}{Komlossy, K.}, \bibinfo{author}{Schuster, S.},
  \bibinfo{author}{Wiegmann, M.}, \bibinfo{author}{Fernandez, E. P.~G.},
  \bibinfo{author}{Hagen, M.}, \& \bibinfo{author}{Stein, B.}
  (\bibinfo{year}{2018}).
\newblock \bibinfo{title}{Crowdsourcing a large corpus of clickbait on
  twitter}.
\newblock In {\it \bibinfo{booktitle}{Proceedings of the 27th {International}
  {Conference} on {Computational} {Linguistics}}\/} (pp.
  \bibinfo{pages}{1498--1507}).
\bibitem[{Potthast et~al.(2016)Potthast, Köpsel, Stein \&
  Hagen}]{potthast_clickbait_2016}
\bibinfo{author}{Potthast, M.}, \bibinfo{author}{Köpsel, S.},
  \bibinfo{author}{Stein, B.}, \& \bibinfo{author}{Hagen, M.}
  (\bibinfo{year}{2016}).
\newblock \bibinfo{title}{Clickbait detection}.
\newblock In {\it \bibinfo{booktitle}{European {Conference} on {Information}
  {Retrieval}}\/} (pp. \bibinfo{pages}{810--817}).
\newblock \bibinfo{publisher}{Springer}.
\bibitem[{Robbani(2018)}]{robbani_github_2018}
\bibinfo{author}{Robbani, H.} (\bibinfo{year}{2018}).
\newblock \bibinfo{title}{Github: {Indonesian} stemmer. {Python} port of {PHP}
  {Sastrawi} project.}
\newblock \URLprefix \url{https://github.com/har07/PySastrawi}.
\bibitem[{William \& Sari(2020)}]{william_click-id_2020}
\bibinfo{author}{William, A.}, \& \bibinfo{author}{Sari, Y.}
  (\bibinfo{year}{2020}).
\newblock \bibinfo{title}{{CLICK}-{ID}: {A} novel dataset for {Indonesian}
  clickbait headlines}.
\newblock {\it \bibinfo{journal}{Data in Brief}\/},  {\it
  \bibinfo{volume}{32}\/}, \bibinfo{pages}{106231}. \URLprefix
  \url{https://linkinghub.elsevier.com/retrieve/pii/S2352340920311252}.
  \DOIprefix\doi{10.1016/j.dib.2020.106231}.
\bibitem[{Wolf et~al.(2019)Wolf, Debut, Sanh, Chaumond, Delangue, Moi, Cistac,
  Rault, Louf \& Funtowicz}]{wolf_transformers_2019}
\bibinfo{author}{Wolf, T.}, \bibinfo{author}{Debut, L.}, \bibinfo{author}{Sanh,
  V.}, \bibinfo{author}{Chaumond, J.}, \bibinfo{author}{Delangue, C.},
  \bibinfo{author}{Moi, A.}, \bibinfo{author}{Cistac, P.},
  \bibinfo{author}{Rault, T.}, \bibinfo{author}{Louf, R.}, \&
  \bibinfo{author}{Funtowicz, M.} (\bibinfo{year}{2019}).
\newblock \bibinfo{title}{Transformers: {State}-of-the-art {Natural} {Language}
  {Processing}}.
\newblock {\it \bibinfo{journal}{arXiv preprint arXiv:1910.03771}\/}, .
\bibitem[{Zuhroh \& Rakhmawati(2020)}]{zuhroh_clickbait_2020}
\bibinfo{author}{Zuhroh, N.~A.}, \& \bibinfo{author}{Rakhmawati, N.~A.}
  (\bibinfo{year}{2020}).
\newblock \bibinfo{title}{Clickbait detection: {A} literature review of the
  methods used}.
\newblock {\it \bibinfo{journal}{Register: Jurnal Ilmiah Teknologi Sistem
  Informasi}\/},  {\it \bibinfo{volume}{6}\/}, \bibinfo{pages}{1--10}.
\newblock \bibinfo{note}{ISBN: 2502-3357}.

\end{thebibliography}

\end{document}